%% file: egpaper.tex
\documentclass[10pt,twocolumn,letterpaper]{article}
\usepackage[accsupp]{axessibility} 
\usepackage{wacv}
\usepackage{times}
\usepackage{epsfig}
\usepackage{graphicx}
\usepackage{amsmath}
\usepackage{amssymb}
\usepackage{nicefrac}       
\usepackage{microtype}      
\usepackage{xcolor}
\usepackage{graphicx}
\usepackage{pifont}
\usepackage{pifont}
\usepackage{enumitem}
\usepackage{mathtools}
\usepackage{siunitx}
\usepackage{booktabs}
\usepackage{multirow}
\usepackage[toc,page,header]{appendix}
\usepackage{minitoc}

\def\wacvPaperID{244} 

\wacvfinalcopy 

\ifwacvfinal
\fi

\ifwacvfinal
\usepackage[breaklinks=true,bookmarks=false]{hyperref}
\else
\usepackage[pagebackref=true,breaklinks=true,colorlinks,bookmarks=false]{hyperref}
\fi

\usepackage{placeins}
\usepackage{wrapfig}
\usepackage[export]{adjustbox}
\newcommand{\cmark}{\ding{51}}%
\newcommand{\xmark}{\ding{55}}%

\title{LEAD: Self-Supervised Landmark Estimation by Aligning Distributions of Feature Similarity}

\author{Tejan Karmali\textsuperscript{1}\thanks{Equal contribution.} \enspace Abhinav Atrishi\textsuperscript{1}\footnotemark[1] \enspace Sai Sree Harsha\textsuperscript{1} \\ Susmit Agrawal\textsuperscript{1} \enspace Varun Jampani\textsuperscript{2} \enspace R. Venkatesh Babu\textsuperscript{1} \\
\small \textsuperscript{1}Indian Institute of Science, Bengaluru \ \ \ \ \ \textsuperscript{2}Google Research}

\begin{document}

\maketitle
\thispagestyle{empty}

\begin{abstract}
    In this work, we introduce LEAD\footnote{Project Page: \href{https://sites.google.com/view/lead-wacv22}{https://sites.google.com/view/lead-wacv22}}, an approach to discover landmarks from an unannotated collection of category-specific images. Existing works in self-supervised landmark detection are based on learning dense (pixel-level) feature representations from an image, which are further used to learn landmarks in a semi-supervised manner. While there have been advances in self-supervised learning of image features for instance-level tasks like classification, these methods do not ensure dense equivariant representations. The property of equivariance is of interest for dense prediction tasks like landmark estimation. In this work, we introduce an approach to enhance the learning of dense equivariant representations in a self-supervised fashion. We follow a two-stage training approach: first, we train a network using the BYOL~\cite{grill2020bootstrap} objective which operates at an instance level. The correspondences obtained through this network are further used to train a dense and compact representation of the image using a lightweight network. We show that having such a prior in the feature extractor helps in landmark detection, even under drastically limited number of annotations while also improving generalization across scale variations.
\end{abstract}

\section{Introduction}
Image landmarks are distinct locations in an image that can provide useful information about the object, like its shape and pose. They can be used to predict camera pose using Structure-from-Motion \cite{hartley_zisserman_2004}. Landmark detection is a well studied problem in computer vision ~\cite{Zhang2016LearningDR, rar, fld, 8578383, wing, Li2019TaskRN} that was initially accomplished using annotated data. Landmark annotation requires a person to accurately label the pixel location where the landmark is present. This makes annotation a laborious, biased, and ambiguous task, motivating the need for newer paradigms such as few-shot learning~\cite{Zhao2018DynamicCN,zhang2017learning, Su2020When, Wang2017ConditionalDG} and self-supervised learning~\cite{NIPS2014_07563a3f, zhang2016colorful, misra2020pirl, He_2020_CVPR, grill2020bootstrap}.  

Prior works in self-supervised landmark detection rely on the principles of reconstruction~\cite{8578383, 8953229} and equivariance~\cite{Thewlis19a, Thewlis_2017_ICCV}. These methods are trained using dense objectives that are satisfied by every pixel (or by every patch of pixels, due to downsampling). This tends to capture only local information around each pixel, and is unaffected by structural changes in the image (like patch shuffling). 

Most of the existing research in the field of self-supervised learning is focused towards the task of instance-level classification. Amongst the proposed pretext tasks for self-supervision, instance-discriminative methods~\cite{He_2020_CVPR, chen2020big, chen2020mocov2, chen2021empirical, pmlr-v119-chen20j}, are known to be superior for the purpose of pre-training. Recent methods utilize these objectives for dense prediction tasks as well, where a distinct label is predicted either for every pixel (segmentation, landmark detection) or patch of pixels (detection)~\cite{roh2021scrl, vaader, xie2020propagate, wang2020DenseCL}. The power of contrastive training is leveraged for landmark detection by Cheng et al.~\cite{cheng2020unsupervised} to achieve state-of-the-art performance using Momentum Contrast (MoCo)-style~\cite{He_2020_CVPR} pre-training. This work demonstrates equivariant properties in the network when trained with a contrastive objective. This property is realised by extracting a hypercolumn-style feature map from the image. But using such a high-dimensional feature map (3840d for ResNet50 due to stacking up of features), which is $60\times$ larger than existing approaches, to represent an image is not scalable to large images. 

\begin{figure}
  \centering
  \includegraphics[width=0.4\textwidth]{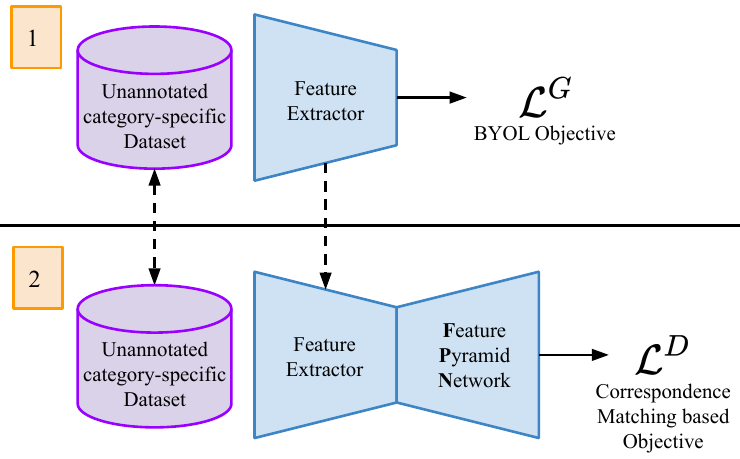}
  \caption{\textbf{LEAD Framework overview. }Two stage process for self-supervised landmark detection. \textbf{First}, an \textit{instance-level} feature extractor is trained on a large Unannotated category-specific Dataset with the BYOL~\cite{grill2020bootstrap} objective. \textbf{Second}, using the correspondence matching property of the instance-level feature extractor, a \textit{pixel-level} FPN~\cite{fpn} based feature extractor is trained on the same dataset. Finally, the pixel-level feature extractor is used to train a supervised regressor on limited data of landmark annotations.}
  \label{fig:overview}
\end{figure}

\begin{figure*}[t]
  \centering
  \includegraphics[width=\textwidth,center]{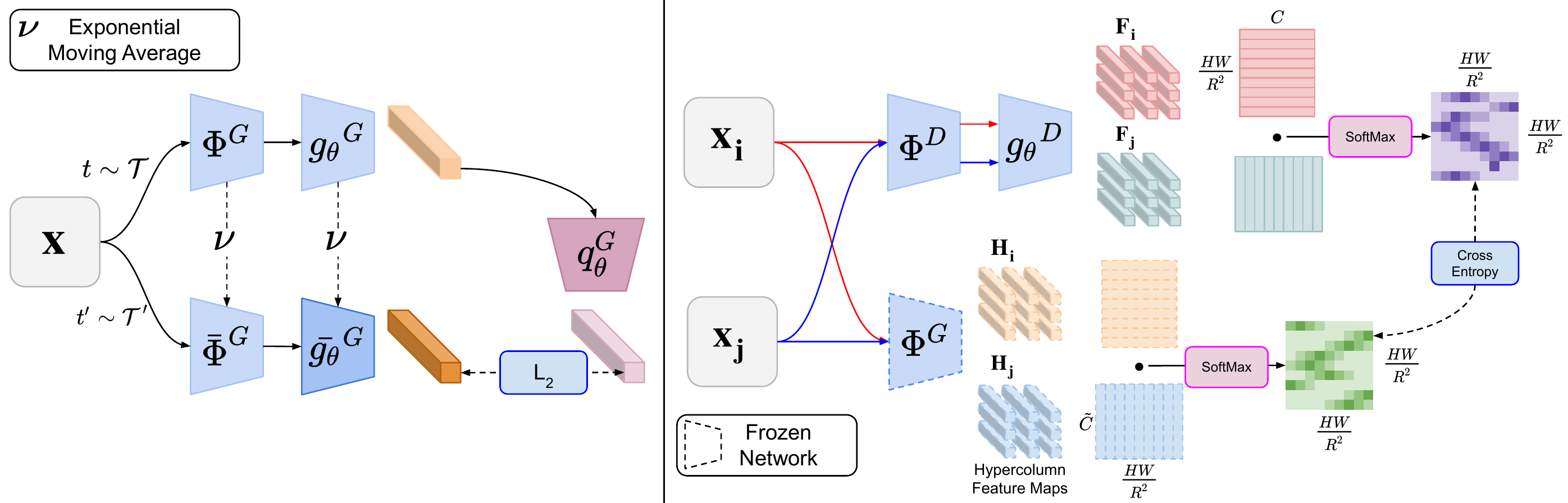}
  \caption{\textbf{LEAD training overview. }\textbf{Left: }Stage 1 of the training feature extractor $\Phi^G$ with BYOL objective, where the representation of key augmentation is predicted from query augmentation. \textbf{Right: }Stage 2 involves using frozen $\Phi^G$ to obtain dense correspondences, which are used to guide trainable network $\Phi^D$ to obtain dense and compact image representation. The correspondences, which also describe similarity between features, are converted to a probability distribution over spatial grid, by using a softmax (ref. Fig.~\ref{fig:tracking}). Distribution of Feature Similarity from $\Phi^D$ is guided by that from $\Phi^G$ using a cross-entropy loss.}
  \label{fig:arch}
\end{figure*}

Our key insight is based on the observation that self-supervised training on category-specific datasets (dataset that consists of images that belong to only single category) leads to meaningful part-clustering in feature space. We further utilize this finding to propose a dense self-supervised objective for landmark prediction. Specifically, LEAD involves two stages: (1) Global representation learning, and (2) Correspondence-guided dense and compact representation learning. The network from stage 1 leads to meaningful part clustering in the feature space, and hence can be used to draw correspondences between two images. This can be used for pixel/patch level training to learn compact descriptors that represent the spatial information of the image. We illustrate the high-level idea in Fig.~\ref{fig:overview}, and include a detailed architecture in Fig.~\ref{fig:arch}.

We measure the performance of LEAD using percentage of inter-ocular distance (IOD). Landmarks estimated using our feature extractor show $\sim$10\% improvement over prior art on facial landmark estimation, along with a boost in performance in the setting of severely limited annotations. We further obtain improved generalization to alignment and scale changes in the input images. 

In summary, our contributions are:
\begin{itemize}
    \item We show the emergence of high-fidelity landmarks in Bootstrap-Your-Own-Latent (BYOL) \cite{grill2020bootstrap} style instance-level feature learning framework. (Sec.~\ref{sec:glob_byol})
    \item We utilise this property to guide the learning of dense and compact feature maps of the image via a novel dimensionality reduction objective. (Sec.~\ref{sec:dense_byol})
    \item Our evaluations show significant improvements over prior art on challenging datasets and across degrees of annotations, both qualitatively and quantitatively. (Sec.~\ref{sec:expts})
\end{itemize}

\section{Related Works}
\noindent\textbf{Unsupervised Landmark prediction:} 
The landmark prediction task has traditionally been studied in a supervised learning setting. Given the annotation-heavy nature of the problem, recent approaches have emphasized on unsupervised pretraining to learn information-rich features. These approaches can be divided based on two principles: equivariance and image generation.

Thewlis et al.~\cite{Thewlis_2017_ICCV} proposed an approach that uses equivariance of the feature descriptors across image warps as an objective for supervision. Suwajanakorn et al.~\cite{NEURIPS2018_24146db4} extended this idea for 3D landmark discovery from multi-view image pairs. This idea has also been used to model symmetrically deformable objects~\cite{NEURIPS2018_1d640826}, and to learn object frames~\cite{NIPS2017_cbcb58ac}. Further, Thewlis et al.~\cite{Thewlis19a} supplemented it using the principle of transitivity, which ensured that the descriptors learnt are robust across images.

Generative objective for landmark detection was initially used by Zhang et al.~\cite{8578383} and Lorenz et al.~\cite{8953229}. The main idea is to learn an image autoencoder with a landmark discovery bottleneck. Jakab et al.~\cite{NEURIPS2018_1f36c15d} coupled it with conditional image generation which could decouple the appearance and pose over an image pair. The key downside of these methods is that, the discovered landmarks are not interpretable. This was addressed by ~\cite{Jakab_2020_CVPR} where the landmark bottleneck is interpretable, due to availability of unpaired poses. ~\cite{unsupervLandm2020} detects more semantically meaningful landmarks using self-training and deep clustering.

\vspace{-1mm}

\noindent\textbf{Self-supervised learning:} 
Self-supervised learning follows the paradigm of training a network using a pretext task on a large-scale unlabeled dataset, followed by training a shallow network using limited annotated data. Initial works explored pretext tasks like classification of image orientation ~\cite{gidaris2018unsupervised}, patch-location prediction ~\cite{Noroozi2016UnsupervisedLO, efros}, image colorization~\cite{zhang2016colorful, zhang2017split}, and clustering ~\cite{Caron_2018_ECCV, asano2020self}. While transformation invariant representation learning of an image~\cite{Zhao2017MarginalizedCL, 10.1145/2964284.2984061, Sohn2012LearningIR, Laptev_2016_CVPR} has been extensively studied in supervised learning, the idea has outperformed prior pretext tasks when modelled as a contrastive learning problem~\cite{He_2020_CVPR, chen2020big, chen2020mocov2, chen2021empirical, pmlr-v119-chen20j} in the self-supervised learning setting. Here, the main idea is to push the embeddings of the \textit{query} image and its augmentation (``positive" image) closer, while repelling it against the embeddings of the ``negative" images (all other images). This is achieved using the InfoNCE~\cite{oord2019representation} loss. A key disadvantage of these methods is the use of a large number of ``negative" images which leads to high memory requirements. This was mitigated by methods like ~\cite{grill2020bootstrap} and ~\cite{chen2020exploring}, which achieve competitive performance without ``negative" images. While both of these seminal works concentrate on the classification task, there are some advances in adapting these techniques for dense prediction tasks like detection and segmentation~\cite{roh2021scrl, vaader, xie2020propagate, wang2020DenseCL} as well. The only work that adopted the contrastive learning objective for the task of landmark prediction is ContrastLandmarks (CL) ~\cite{cheng2020unsupervised}, where they train the network with the InfoNCE~\cite{oord2019representation} objective. To adapt the output feature map to the resolution of the image, they use a hypercolumn representation from features across different layers. The key differences between this work and LEAD are: 1) We learn dense and compact descriptors via a novel correspondence matching guided dimensionality reduction objective while CL uses the objective proposed by Thewlis et al.~\cite{NIPS2017_cbcb58ac}, and 2) We do not use any ``negative" images, as landmarks are ubiquitous in a category-specific dataset.

\section{Method}
\label{sec:method}

\subsection{Background}
Let $\mathcal{X} = \{\boldsymbol{x} \in \mathbb{R}^{H \times W \times 3}\}$ be a large-scale unannotated category-specific dataset. Our goal is to learn a feature extractor $\Phi$, which, given $\boldsymbol{x} \in \mathcal{X}$ as input gives a feature map as output. As a pretext task, prior works have attempted to enforce instance-level representations to be invariant to transformations~\cite{cheng2020unsupervised}, and impose consistency on the dense pixel-level representations. In our approach LEAD, we use two stages. First, we learn a global representation of the image that leads to its part-wise clustering as described in Sec.~\ref{sec:glob_byol}. Then, we make use of this prior to guide the learning of a dense and compact representation of the image by a novel dimensionality reduction objective, which matches the distributions of feature similarity across two images, as described in Sec.~\ref{sec:dense_byol}.


\subsection{Global Representation Learning}
\label{sec:glob_byol}
We follow the algorithm proposed in BYOL~\cite{grill2020bootstrap} to learn an instance-level representation of the image. BYOL uses an online network $\Phi^G$ and a target network $\bar{\Phi}^G$. $\Phi^G$ and $\bar{\Phi}^G$ share the same architecture, but the weights of $\bar{\Phi}^G$ are obtained using a momentum average of weights of ${\Phi}^G$ across multiple training iterations. These backbone networks are followed by projection heads $g_\theta^G$ and $\bar{g_\theta}^G$. Similar to the weights of the backbone, the weights of $\bar{g_\theta}^G$ are obtained using a momentum average. The necessity for the projection heads in self-supervised training has been discussed extensively in SimCLR~\cite{pmlr-v119-chen20j}, where the authors find the representations of last layer before the projection head to be most useful. Additionally, the online network has a prediction head $q_\theta^G$ (Fig.~\ref{fig:arch}).

The training objective is to predict the representation of one view of the image from another using $q_\theta^G$. Given an image $\boldsymbol{x}$, its two views $\boldsymbol{x_1}$ and $\boldsymbol{x_2}$ are generated by applying augmentations. We refer to $\boldsymbol{x_1}$ as the \textit{query image} and $\boldsymbol{x_2}$ as the \textit{key image}. $\Phi^G$ and $\bar{\Phi}^G$ generate features corresponding to $\boldsymbol{x_1}$ and $\boldsymbol{x_2}$ respectively. These feature maps are then projected using $g_\theta^G$ and $\bar{g_\theta}^G$ respectively to obtain the instance-level representations $\boldsymbol{z_1}$ and $\boldsymbol{z_2}$. Since both the views belong to the same instance, the predictor $q_\theta^G$ is trained to predict $\boldsymbol{z_2}$ given $\boldsymbol{z_1}$. The squared L$_2$ loss shown below is minimized for training:
\begin{equation}
  \mathcal{L}^G = \| q_{\theta}^G(\boldsymbol{z_1}) - \boldsymbol{z_2}\|_2^2
  \label{eq:dense}
\end{equation}
\begin{figure}
    \centering
    \includegraphics[width=0.5\textwidth]{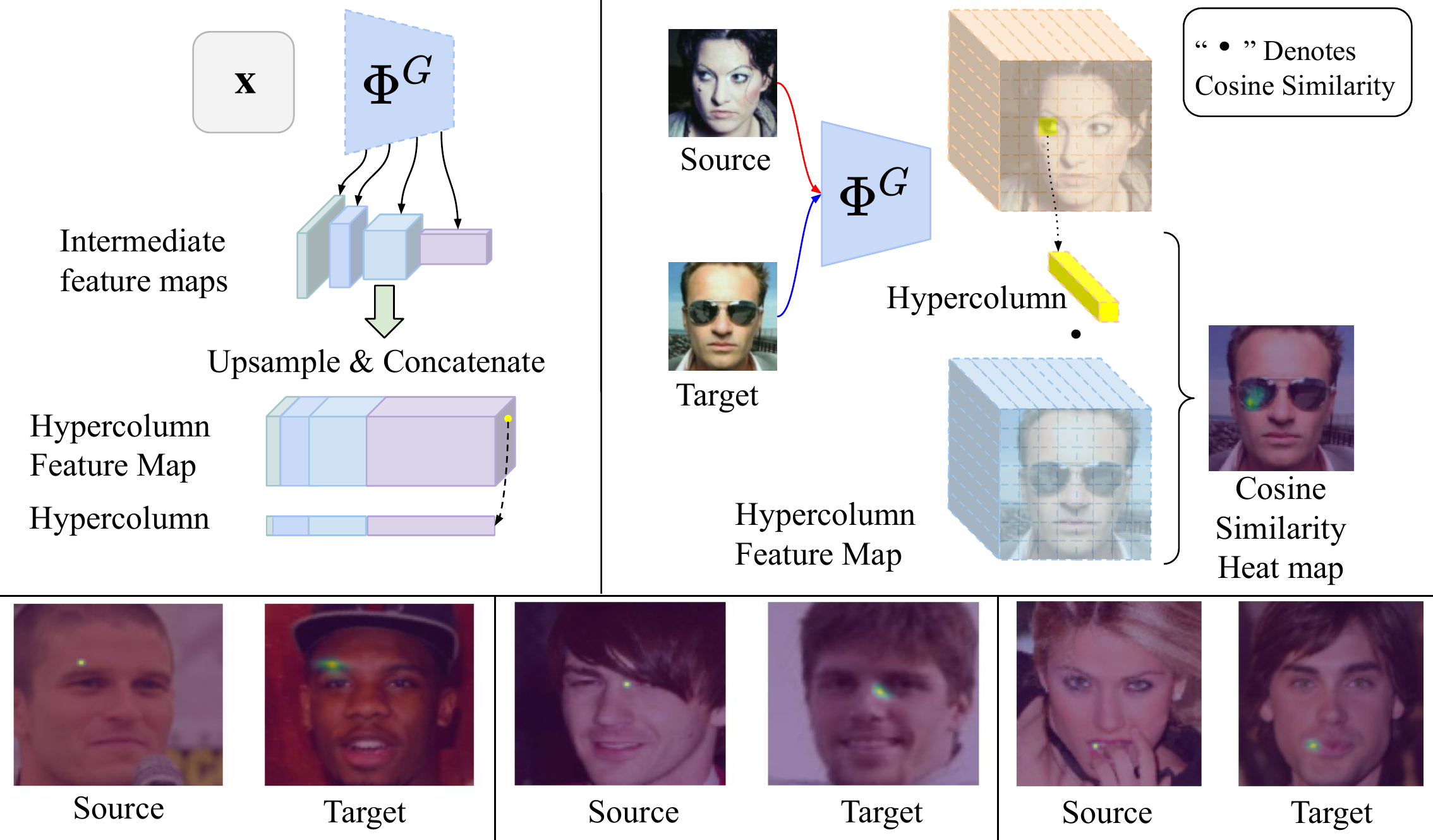}
    \caption{\textbf{Correspondence matching} performance using the hypercolumn representation. \textbf{Top Left: } Procedure to create hypercolumn from intermediate feature maps, by upsampling and concatenating them. Each feature vector across the spatial dimension denotes a hypercolumn. \textbf{Top Right: } Correspondence matching is performed from a point in the source image to the target image by taking cosine similarity of the hypercolumn corresponding to the source point and target's hypercolumn feature map, followed by softmax to obtain a heat map. \textbf{Bottom: }Examples of correspondence matching. Note that the resultant distribution peaks around the tracked point.}
    \label{fig:tracking}
    \vspace{-6mm}
\end{figure}

As shown in CL~\cite{cheng2020unsupervised}, the self-supervised contrastive objective produces hypercolumn based feature maps that have semantic understanding of the correspondences at pixel level between two images. In addition, we find that the BYOL objective gives significantly better correspondences than the MoCo objective, as shown in the Fig.~\ref{fig:tracking}. Hypercolumns are used here, since the self-supervised networks downsample the input image largely to obtain an instance-level representation. Creating a hypercolumn based feature map involves concatenating the intermediate feature maps along the channel dimension. Since the intermediate feature maps have lower spatial resolution than the original input image, they are upsampled to match the resolution of the input image. This has been illustrated in Fig.~\ref{fig:tracking}. However, hypercolumns incur a large cost in terms of memory. In the next section, we improve upon this by injecting pixel-level information into the network, thereby learning a dense and compact representation of the image.

\subsection{Dense and Compact Representation Learning}
\label{sec:dense_byol}

The bottleneck in framing the dense feature map learning problem is pixel-level correspondences. In the case of global feature vector learning, the image to form the positive pair is drawn by applying augmentation to the input image. But in the case of dense feature map learning, the correspondences between points in the query and the key images are not known. But since we have a trained BYOL network that can find \textit{reasonable} (ref. Fig.~\ref{fig:tracking}) correspondences across images, we use it to guide the learning of dense and compact feature maps of images. 

For the hypercolumn feature vector (or hypercolumn, for short), the ability to track a semantic point across two image depends on the distance between them in the $\tilde{C}$-d feature space. In this space, the features are clustered according to their semantic meaning. We aim to learn a compact feature space which has this property.

We now elaborate on the training method followed to learn such a low-dimensional feature space (Fig.~\ref{fig:arch}). We train an encoder-decoder network $\Phi^D: \mathbb{R}^{H \times W \times 3} \to \mathbb{R}^{\frac{H}{R} \times \frac{W}{R} \times C}$. The encoder is initialised with $\Phi^G$ trained in Sec.~\ref{sec:glob_byol}. The output of the encoder goes to the projection head $g_\theta^D$. We aim to retain the relationship defined by the cosine similarity between the hypercolumn feature maps from two images in their compact feature maps which are to be learnt. Let $\boldsymbol{x}_i$, $\boldsymbol{x}_j \in \mathcal{X}$ be two images, whose hypercolumn feature maps are $\boldsymbol{H}_i$, $\boldsymbol{H}_j \in \mathbb{R}^{H \times W \times 3} \to \mathbb{R}^{\frac{H}{R} \times \frac{W}{R} \times \tilde{C}} $ respectively. Note that, $\tilde{C} \gg C$, which makes the hypercolumn representation memory-intensive during inference. Let $\boldsymbol{F}_i = g_{\theta}^D(\Phi^D(\boldsymbol{x}_i))$ be compact feature maps of the respective images. Let $\boldsymbol{f}_i^{uv} \in \boldsymbol{F}_i$ be a feature vector at spatial location $(u, v)$ in feature map $\boldsymbol{F}_i$. Similarly, let $\boldsymbol{h}_i^{uv} \in \boldsymbol{H}_i$ be a feature vector at spatial location $(u, v)$ in the hypercolumn feature map $\boldsymbol{H}_i$. Since the aim is to retain the inter-feature relationship, we use cosine similarity as the measure of relationship between two feature vectors. To cover the whole feature space, we take cosine similarity with all the feature vectors. This relationship between the feature vector and the feature space as a probability distribution indicates which subspace of the feature space the feature vector is most similar to:
\begin{equation}
    q_{ij}^{uv}[k, l] = \frac{\exp({\boldsymbol{f}_i^{uv}}^T \boldsymbol{f}_j^{kl} / \tau)}{\sum_{m,n=0}^{\frac{H}{R}, \frac{W}{R}}\exp({\boldsymbol{f}_i^{uv}}^T \boldsymbol{f}_j^{mn}/ \tau)}
    \label{eq:dist}
\end{equation}
where $\tau$ is temperature, which is a hyperparameter controlling the concentration level of the probability distribution $q_{ij}^{uv}$ ~\cite{Wu_2018_CVPR}.

Similarly, such a relationship can be defined for $\boldsymbol{h}_i^{uv}$ with $\boldsymbol{H}_j$ as well. We denote this probability distribution as $p_{ij}^{uv}$. This ultimately leads us to optimize $q_{ij}^{uv}$ to mimic $p_{ij}^{uv}$. We use cross-entropy between the both of them to achieve this objective:
\vspace{-2mm}
\begin{equation}
    \mathcal{L}^D = \sum_{u,v=0}^{\frac{H}{R}, \frac{W}{R}} \sum_{k,l=0}^{\frac{H}{R}, \frac{W}{R}}-p_{ij}^{uv}[k, l] \cdot \log(q_{ij}^{uv}[k, l])
\end{equation}


\subsection{Landmark Detection}
At this stage we have a feature extractor that is learned in a self-supervised fashion. To obtain the final landmark prediction, a limited amount of annotated data is used. Feature extractor is frozen and a lightweight predictor $\Psi$ is trained over it. $\Psi$ gives landmark heatmaps as output ($\in \mathbb{R}^{H \times W \times K}$) where $K$ is the number of landmarks present). Expected location of the landmark $k$, weighed by the heatmap gives its final position $(\hat{x}^k, \hat{y}^k)$. It is supervised by the annotated location of the landmark $(x^k, y^k)$ with an $l_2$ loss.

\section{Experiments}
\label{sec:expts}
\textbf{Dataset: } We evaluate LEAD on human faces. Following prior works, we use the CelebA~\cite{liu2015faceattributes} dataset containing 162,770 images for pretraining the network. To evaluate the learnt representation, four datasets are used. We firstly use MAFL which is a subset of CelebA. Two variants of AFLW~\cite{aflw} are used: the first being AFLW$_{M}$ which is the partition of AFLW with crops from MTFL~\cite{fld}. It contains 10,122 training images and 2,995 test images. The second variant is AFLW$_{R}$, in which tighter crops of the face are used. This comprises of 10,122 training images and 2,991 testing images. We further use the 300-W ~\cite{300w} dataset which has 68 annotated landmarks, with 3148 training and 689 testing images. All the datasets are publicly available.

\begin{figure*}
    \centering
    \includegraphics[width=\textwidth]{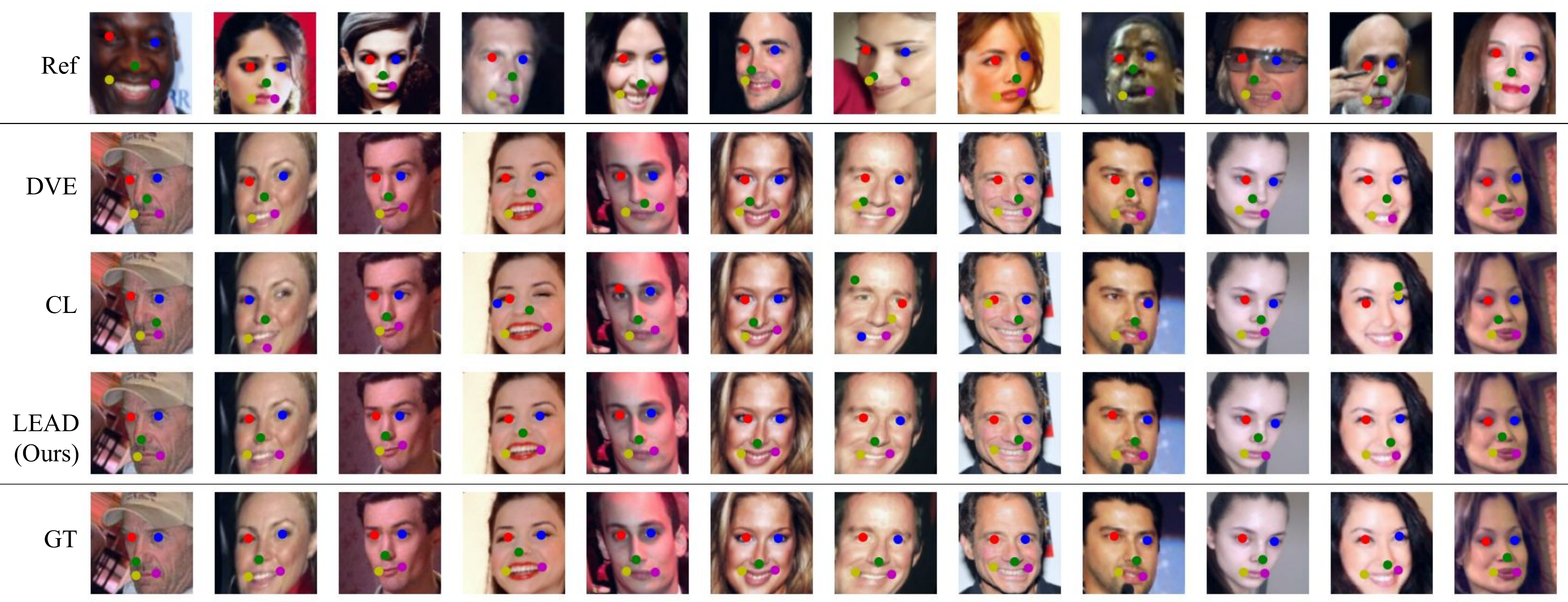}
    \caption{\textbf{Landmark Matching: } We observe that LEAD is able to predict the landmarks in the Query images (middle rows) using reference annotated image (first row). We compare our performance against DVE~\cite{Thewlis19a} and ContrastLandmarks~\cite{cheng2020unsupervised} on a spectrum of head rotations.}
    \label{fig:matching}
    \vspace{-3mm}
\end{figure*}

\textbf{Implementation Details: } We use a ResNet50~\cite{resnet50} backbone to train instance-level BYOL representation in stage 1. In stage 2, the feature extractor of the trained ResNet in stage 1 is used as weight initialization for the encoder. The decoder is made up of FPN~\cite{fpn}. It is a lightweight network following the encoder which incorporates features from multiple scales of the encoder to create the final dense feature representation. This idea is similar to the creation of a hypercolumn feature map. FPN builds the final representation from features at 1/4, 1/8, 1/16 and 1/32 scales, using upsampling blocks as proposed in~\cite{vaader}. The final dense representation has a feature dimension of 64 and spatial downscaling of 1/4. The feature projection head is composed of 2 linear layers with BatchNorm and ReLU.

We use BYOL for stage 1 training with a batch size of 256 for 200 epochs using the SGD optimizer. The learning rate is set to \num{3e-2} with a cosine decay for stage 1 training. For stage 2, we train with a batch size of 256 for 20 epochs on the CelebA dataset. We set the temperature $\tau$ to be 0.05. For a fair comparison we train the supervised regressors with frozen feature extractor as proposed in~\cite{cheng2020unsupervised}. The regressor initially comprises of $50$ filters (to keep evaluations consistent with \cite{cheng2020unsupervised, Thewlis19a}) of dimension $1 \times 1 \times K$ which transforms the input feature maps to heatmaps of intermediate virtual keypoints. These heatmaps are converted to $2K$ x-y pairs using a $softargmax$ layer,  which are further linearly regressed to estimate manually annotated landmarks. Here $K$ represents the number of annotated keypoints in the dataset. Following DVE, we resize the input image to ($136 \times 136$) and then take a ($96 \times 96$) central crop for performing the evaluations. For stage 1 training we take two ($96 \times 96$) sized random crops. We perform all of our experiments on 2 Tesla V100 GPUs.
\vspace{-0.5mm}

\textbf{Evaluation: }Following prior works, we use percentage of inter-ocular distance (IOD) as the error. We evaluate on two tasks, landmark matching and landmark regression. We describe each of the evaluation tasks next. 

\subsection{Landmark Matching}

\begin{table}
{\centering
\caption{\textbf{Landmark matching} performance comparison against prior art on MAFL dataset. The error is reported as a percentage of inter-ocular distance. }
\label{tab:matching}
\begin{tabular}{ll|ll}
\toprule
\multicolumn{1}{l}{\textbf{Method}} & \multicolumn{1}{c|}{\textbf{Feat. dim.}} & \multicolumn{1}{c}{\textbf{Same}} & \multicolumn{1}{c}{\textbf{Different}} \\ \midrule
DVE~\cite{Thewlis19a}               & 64                                        & 0.92                              & \textbf{2.38}                                   \\
CL~\cite{cheng2020unsupervised}& 64                              & 0.92                              & 2.62                                   \\
BYOL + NMF                               & 64                                        & 0.84                               & 5.74                                   \\
LEAD (ours)                                & 64                                        & \textbf{0.51}                               & \underline{2.60}                                   \\  \midrule
CL~\cite{cheng2020unsupervised}& 256                             & 0.71                              & \textbf{2.06}                                   \\
BYOL + NMF                                & 256                                       & 0.91                               & 4.26                                    \\
LEAD (ours)                                & 256                                       & \textbf{0.48}                               & \underline{2.50}                                    \\  \midrule
CL~\cite{cheng2020unsupervised}& 3840                            & 0.73                              & 6.16                                   \\
LEAD (ours)                                & 3840                                      & \textbf{0.49}                               & \textbf{3.06}               \\ \bottomrule                   
\end{tabular}
}
\vspace{-5mm}
\end{table}

In the landmark matching task, we are given two images. One is a reference image for which the landmarks are known and the other is a query image, for which the landmarks are to be predicted. Prediction is done by choosing the feature descriptor of a landmark in the reference image, and finding the location of the most similar feature descriptor to it in the feature map of the query image using cosine similarity. In line with DVE \cite{Thewlis19a}, we evaluate on a dataset consisting of 500 same identity and 500 different identity pairs taken from MAFL. Qualitative results of matching are shown in Fig.~\ref{fig:matching}, while quantitative results are presented in Table \ref{tab:matching}. Also shown in Table \ref{tab:matching} is the Non-negative Matrix Factorization (NMF~\cite{lee1999learning}, which gives low-rank approximation of non-negative matrix) baseline, wherein we apply NMF over the learned hypercolumn thereby showing that our dimensionality reduction objective is superior to naively applying NMF over the learned hypercolumn. Similar to the trends from correspondence matching using hypercolumn in Fig.~\ref{fig:tracking}, the final dense model with 64 dimensional features is able to meaningfully match the landmarks from reference image to query image. This is verified across a head rotation ranging from left-facing to frontal faces and right-facing images. The matching is consistent across genders, showing no bias for any gender.

\vspace{-2mm}

\subsection{Landmark Regression}

\begin{table}[t]
\centering
\caption{\textbf{Landmark regression} performance comparison against prior art. The error is reported as a percentage of inter-ocular distance. We achieve state-of-the-art result on the challenging AFLW datasets with $\sim$10\% relative gain, while obtaining competitive results on MAFL and 300W.}
    \label{tab:regress}
    \resizebox{\linewidth}{!}{%
    \begin{tabular}{lc|cccc}
        \toprule
        \textbf{Method}                                 & \textbf{Unsupervised} & \textbf{MAFL} & \textbf{AFLW}$_M$ & \textbf{AFLW}$_R$ & \textbf{300W} \\     \midrule
        TCDCN~\cite{Zhang2016LearningDR}                                                 & \xmark                & 7.95        & 7.65          & -             & 5.54          \\
        RAR~\cite{rar}                                                   & \xmark                & -           & 7.23          & -             & 4.94          \\
        MTCNN~\cite{fld, 8578383}                                                 & \xmark                & 5.39        & 6.90          & -             & -             \\
        Wing Loss~\cite{wing}                                             & \xmark                & -           & -             & -             & 4.04          \\ \midrule
        \textbf{Dense objective based}                          &                       &             &               &               &               \\
        Sparse~\cite{Thewlis_2017_ICCV}                                                & \cmark                & 6.67        & 10.53         & -             & 7.97          \\
        Structural Repr.~\cite{8578383}                                      & \cmark                & 3.15        & -             & 6.58          & -             \\
        FAb-Net~\cite{Wiles18a}                                               & \cmark                & 3.44        & -             & -             & 5.71          \\
        Def. AE~\cite{def_ae}                                               & \cmark                & 5.45        & -             & -             & -             \\
        Cond. Im. Gen~\cite{NEURIPS2018_1f36c15d}                                         & \cmark                & 2.86        & -             & 6.31          & -             \\
        Int. KP.~\cite{Jakab_2020_CVPR}                                              & \cmark                & -           & -             & -             & 5.12          \\
        Dense3D~\cite{NIPS2017_cbcb58ac}                                               & \cmark                & 4.02        & 10.99         & 10.14         & 8.23          \\
        DVE SmallNet~\cite{Thewlis19a}                                       & \cmark                & 3.42        & 8.60          & 7.79          & 5.75          \\
        DVE Hourglass~\cite{Thewlis19a}                                       & \cmark                & 2.86        & 7.53          & 6.54          & \textbf{4.65} \\ \midrule
        \textbf{Global Objective based}                         &                       &             &               &               &               \\
        ContrastLandmarks~\cite{cheng2020unsupervised}                     & \cmark                & \underline{2.44}& 6.99         & 6.27          & 5.22          \\ 
        LEAD (ours)                                   & \cmark                &\textbf{2.39}&\textbf{6.23}& \textbf{5.65}&\underline{4.66}\\ \bottomrule
    \end{tabular}}
    \vspace{-5mm}
\end{table}

\begin{figure*}
    \includegraphics[width=\textwidth,center]{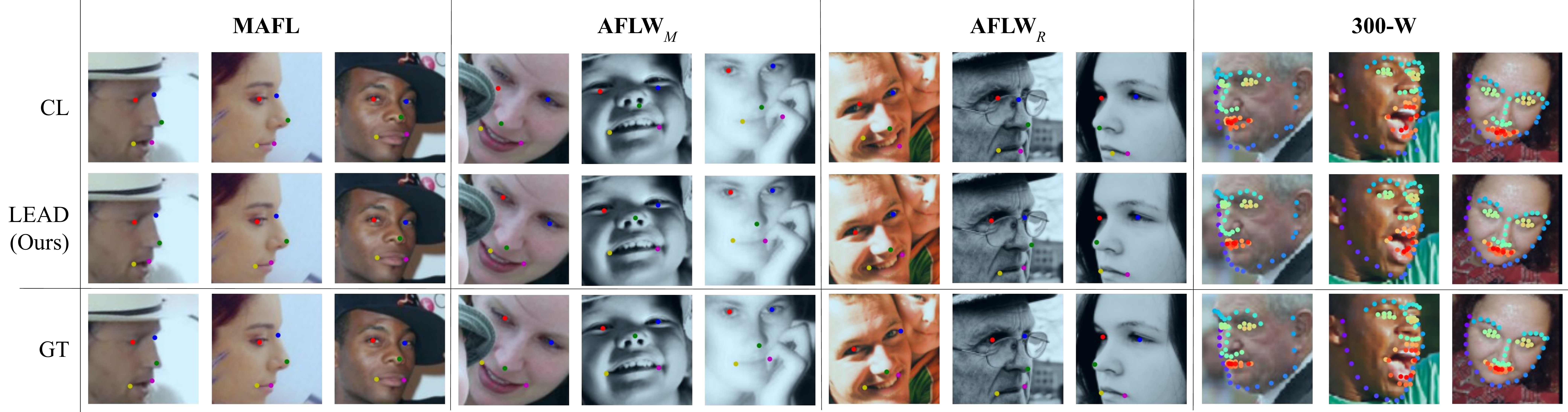}
    \caption{\textbf{Landmark regression:} We observe that features generated by pretraining using LEAD can easily be used to train a lightweight regressor to predict landmarks with high precision. Furthermore, the model is robust to aspects such as face orientation, lighting and minor occlusions.}
    \label{fig:regression}
\end{figure*}

In the task of landmark regression, a lightweight regressor is trained on top of the features extracted by the pretrained network. This is done using supervised learning on the evaluation dataset. We report the inter-ocular distance on landmark regression in Table~\ref{tab:regress}. Our model trained using the BYOL objective achieves results which are  $\sim$10\% better than the prior-art on a relative scale, on 2 out of 4 evaluation datasets, while maintaining a competitive performance on the 300-W dataset. Regression performance is qualitatively verified in the Fig.~\ref{fig:regression}. We refer the reader to the supplementary material for additional datasets and visualizations.

\subsection{Interpretability}

\begin{figure}
    \centering
    \includegraphics[width=0.5\textwidth]{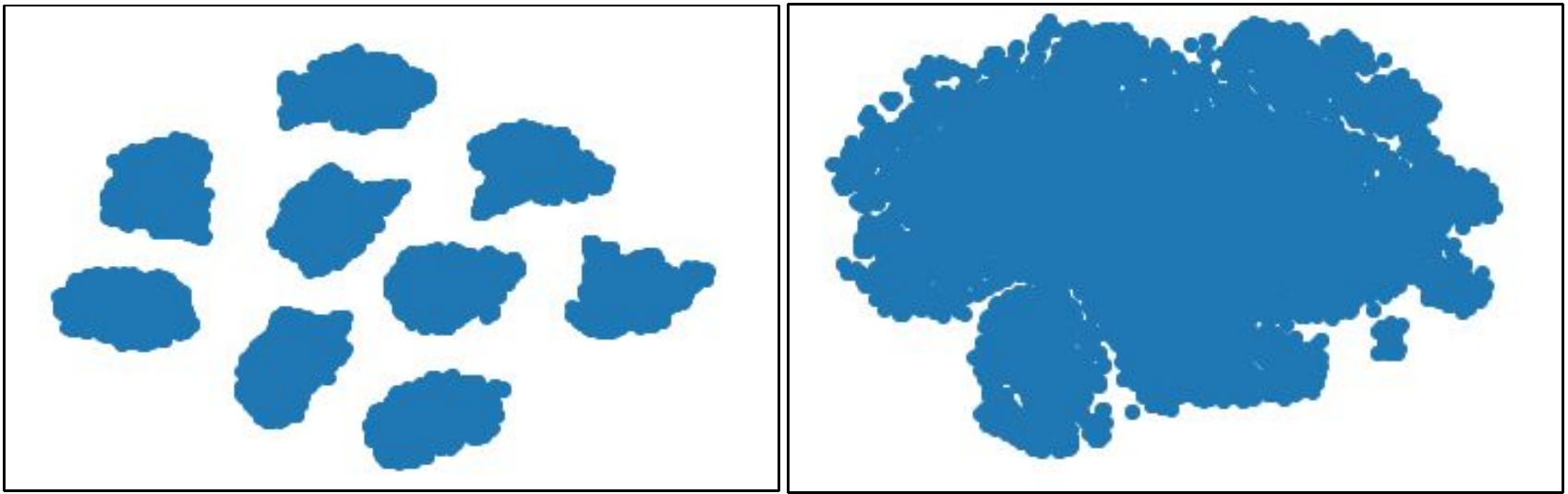}
    \caption{\textbf{t-SNE plots of output feature maps.} \textbf{Left:} LEAD stage 1 features \textbf{Right:} CL stage 1 features}
    \label{fig:byol_clusters}
\end{figure}

\begin{figure}
    \centering
    \includegraphics[width=0.5\textwidth]{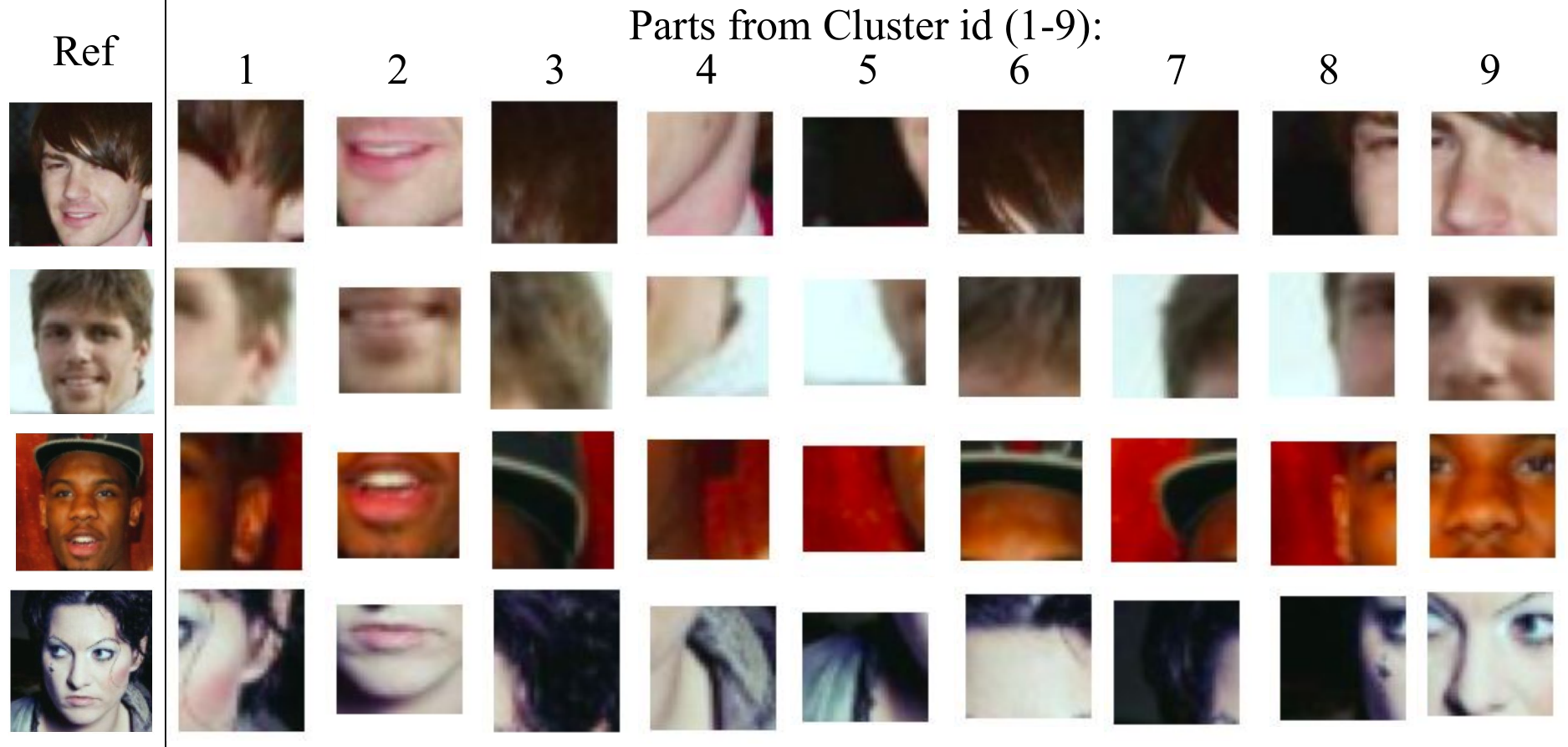}
    \caption{\textbf{t-SNE embeddings tend to cluster part-wise}. The 9 parts (along row) shown for each reference figure (along column) here belong to the 9 clusters in Fig.~\ref{fig:byol_clusters}. Each cluster denotes a semantic part of the face.}
    \label{fig:tsne}
    \vspace{-5mm}
\end{figure}

We observe that the t-SNE embeddings obtained from our model trained with BYOL objective are interpretable. It divides the face spatially into 9 parts, where each clusters corresponds to one of the 9 parts. t-SNE clustering is visualized in Fig.~\ref{fig:byol_clusters} and interpretability of the clusters is verified in Fig.~\ref{fig:tsne}. We also compare our t-SNE plots against that of CL~\cite{cheng2020unsupervised}, wherein we see that CL embeddings are not well clustered when compared to LEAD which shows distinct clusters. We provide further feature clustering analysis in the supplementary material.




\subsection{Ablation Studies}


\begin{table*}
\parbox{.45\linewidth}{
\centering
\caption{\textbf{Effect of projection head on landmark matching}. Projection head affects the matching on different identity. On increasing the dimension of the projection head's output, improvement is observed. Further gains are observed on increasing the final representation's ($\Phi^D$'s output) dimension.}
\label{tab:proj}
\begin{tabular}{ll|ll}
\toprule
\multicolumn{1}{c}{\textbf{Feat. dim.}} & \textbf{Proj. dim.} & \multicolumn{1}{c}{\textbf{Same}} & \multicolumn{1}{c}{\textbf{Different}} \\ \midrule
64                                           & \xmark                        & \textbf{0.48}                              & 2.79                                   \\
64                                           & 64                            & 0.51                              & 2.64                                   \\
64                                           & 256                           & 0.51                              & \textbf{2.60}                                   \\ \midrule
128                                          & 256                           & 0.47                              & 2.58                                   \\ \midrule
256                                          & 256                           & \textbf{0.48}                              & \textbf{2.50}                                   \\ \bottomrule
\end{tabular}
}
\hfill
\parbox{0.6\linewidth}{
\centering
\caption{Effect of feature dimension on landmark regression task}
\label{tab:feat_dim}
\begin{tabular}{ll|llll}
\toprule
\multicolumn{1}{l}{\textbf{Method}} & \textbf{Feat. dim.} & \multicolumn{1}{c}{\textbf{MAFL}} & \multicolumn{1}{c}{\textbf{AFLW}$_M$} & \textbf{AFLW}$_R$ & \textbf{300W} \\ \midrule
DVE~\cite{Thewlis19a}                                 & 64                 & 2.86                              & 7.53                              & 6.54          & \textbf{4.65} \\
CL~\cite{cheng2020unsupervised}                                  & 64                 & \textbf{2.77}                     & 7.21                              & \textbf{6.22} & 5.19          \\
LEAD (ours)                                & 64                & 2.93                              & \textbf{6.61}                     & 6.32          & 5.32          \\ \midrule
CL~\cite{cheng2020unsupervised}                                  & 128                & \textbf{2.71}
& 7.14                              & \textbf{6.14} & \textbf{5.09} \\
LEAD (ours)                                & 128                & 2.91                              & \textbf{6.60}                     & 6.21          & 5.41          \\ \midrule
CL~\cite{cheng2020unsupervised}                                  & 256                & \textbf{2.64}
& 7.17                              & 6.14 & \textbf{4.99} \\
LEAD (ours)                                & 256                & 2.87                              & \textbf{6.51}                     & \textbf{6.12}          & 5.37          \\ \bottomrule
\end{tabular}
}
\end{table*}

\begin{table*}[h]
\centering
 \caption{\textbf{Number of annotations: } LEAD consistently produces the lowest inter-ocular distance under the presence of different levels of annotations on the AFLW$_M$ dataset. The relative improvement is as high as \textbf{45\%} over previous best (in case of `5 annotations' training setting)}
  \label{tab:fewshot}
    \begin{tabular}{ll|cccccc}
    \toprule
    \multirow{2}{*}{\textbf{Method}} & \multirow{2}{*}{\textbf{Feat. dim.}} & \multicolumn{6}{c}{\textbf{Number of annotations}} \\ \cline{3-8} & & \multicolumn{1}{c}{\textbf{1}} & \multicolumn{1}{c}{\textbf{5}} & \multicolumn{1}{c}{\textbf{10}} & \multicolumn{1}{c}{\textbf{20}} & \multicolumn{1}{c}{\textbf{50}} & \multicolumn{1}{c}{\textbf{100}} \\ \midrule
    DVE~\cite{Thewlis19a}             & 64                  & \textbf{14.23 ± 1.45}          & \textbf{12.04 ± 2.03}          & 12.25 ± 2.42                    & 11.46 ± 0.83                    & 12.76 ± 0.53                    & 11.88 ± 0.16                     \\
    CL~\cite{cheng2020unsupervised}              & 64                  & 24.87 ± 2.67                   & 15.15 ± 0.53                   & 13.62 ± 1.08                    & 11.77 ± 0.68                    & 11.57 ± 0.03                    & 10.06 ± 0.45                     \\
    LEAD (Ours)     & 64                  & 21.8 ± 2.54                    & 13.34 ± 0.43                   & \textbf{11.50 ± 0.34}           & \textbf{10.13 ± 0.45}           & \textbf{9.29 ± 0.45}            & \textbf{9.11 ± 0.25}             \\ \midrule
    CL~\cite{cheng2020unsupervised}              & 128                 & 27.31 ± 1.39                   & 18.66 ± 4.59                   & 13.39 ± 0.30                    & 11.77 ± 0.85                    & 10.25 ± 0.22                    & 9.46 ± 0.05                      \\
    LEAD (ours)     & 128                 & \textbf{21.20 ± 1.67}          & \textbf{13.22 ± 1.43}          & \textbf{10.83 ± 0.65}           & \textbf{9.69 ± 0.41}            & \textbf{8.89 ± 0.2}             & \textbf{8.83 ± 0.33}             \\ \midrule
    CL~\cite{cheng2020unsupervised}              & 256                 & 28.00 ± 1.39                   & 15.85 ± 0.86                   & 12.98 ± 0.16                    & 11.18 ± 0.19                    & 9.56 ± 0.44                     & 9.30 ± 0.20                      \\
    LEAD (ours)     & 256                 & \textbf{21.39 ± 0.74}          & \textbf{12.38 ± 1.28}          & \textbf{11.01 ± 0.48}           & \textbf{10.06 ± 0.59}           & \textbf{8.51 ± 0.09}            & \textbf{8.56 ± 0.21}             \\ \midrule
    CL~\cite{cheng2020unsupervised}              & 3840                & 42.69 ± 5.10                   & 25.74 ± 2.33                   & 17.61 ± 0.75                    & 13.35 ± 0.33                    & 10.67 ± 0.35                    & 9.24 ± 0.35                      \\
    LEAD (ours)     & 3840                & \textbf{24.41 ± 1.38}          & \textbf{14.11 ± 1.30}          & \textbf{11.45 ± 0.88}           & \textbf{10.21 ± 0.44}           & \textbf{8.43 ± 0.25}            & \textbf{8.09 ± 0.28}             \\ \bottomrule
\end{tabular}
\end{table*}

We ablate LEAD on factors like feature dimension, contribution from each stage, projection head, degree of annotation availability, and sensitivity to scale variations.

\noindent\textbf{Feature Dimensions.}
Feature dimension plays a significant role in the landmark regression task. Since the regressor takes features as input, its capacity depends on the dimensions of the feature, i.e. a higher dimensional feature implies that the regressor has more capacity to learn, resulting in better predictions. Our experiments in Table~\ref{tab:feat_dim} indicate a superior performance on the challenging AFLW$_M$ dataset, while achieving competitive performance on MAFL and AFLW$_R$. Surprisingly, we find a large deviation in performance trends on the 300W dataset compared to the results obtained using hypercolumn feature maps (ref. Tab. ~\ref{tab:regress}) as guidance for the compact feature maps.

\noindent\textbf{How much does stage 2 objective contribute?}
To answer this question, we run experiments on 2 different  pretraining (stage 1) objectives, followed by 2 different dimensionality reduction (stage 2) objectives. To compare directly, we take CL's~\cite{cheng2020unsupervised} pretraining and dimensionality reduction objectives and our objectives for the same. We keep the architectures same as LEAD and only vary the objective function for a fair comparison. We report our findings in Table~\ref{tab:dim_red_obj}. Irrespective of the stage 1 training, LEAD's dimensionality reduction procedure improves the IOD.
\begin{table}[]
\centering
\caption{\textbf{Dimensionality reduction objective.} LEAD's proposed dimensionality reduction objective significantly improves the performance irrespective of the global representation learning objective. Results are reported on AFLW$_M$ dataset.}
\label{tab:dim_red_obj}
\begin{tabular}{cc|ccc}
\toprule
\textbf{Global  Rep. Obj.}             & \textbf{Dim. Red. Obj.}            & \multicolumn{3}{c}{\textbf{Feat. dim.}}                   \\ \cline{3-5} 
\multicolumn{1}{c}{\textbf{(Stage 1)}} & \multicolumn{1}{c|}{\textbf{(Stage 2)}} & \textbf{64}                 & \textbf{128}                & \textbf{256}                \\ \midrule
CL                                     & CL                                      & 7.86                        & 7.81                        & 7.31                        \\
CL                                     & LEAD                                    & \textbf{6.66}               & \textbf{6.58}               & \textbf{6.69}               \\ \midrule
LEAD                                   & CL                                      & 7.89 & 7.86 & 7.41 \\
LEAD                                   & LEAD                                    & \textbf{6.61}               & \textbf{6.60}               & \textbf{6.51}               \\ \bottomrule
\end{tabular}
\vspace{-5mm}
\end{table}

\noindent\textbf{Is the projection head necessary in stage 2?} Necessity of the projection head in self-supervised learning has been empirically shown to lead to meaningful representations~\cite{pmlr-v119-chen20j}. We use it in our stage 1 training. However in stage 2, where we aim to get higher resolution feature maps as output, is the projection head still required? We use a projection head $g_\theta^D$ on the final feature map as given by $\Phi^D$ to apply the loss during training. Eventually the $g_\theta^D$ is discarded and only $\Phi^D$ is utilised. Here, we ablate on the performance shown by $\Phi^D$ in the absence of projection head as well as the on the output dimension of the $g_\theta^D$. Since we discard $g_\theta^D$, we are allowed to keep its output's dimension as high as required. In our ablation (ref. Table~\ref{tab:proj}), it is observed that for landmark matching on the same identity, there are marginal changes upon having $g_\theta^D$ as well as varying its output dimension. But the projection head emerges as a distinguishing component in case of matching on different identity. Consistent improvements are observed on increasing the feature dimension of the projection head. It can be seen that this leads to slight degradation of performance on the same identity. We also observe the effect of increasing the feature dimension by keeping the projection dimension fixed where we note a further improvement on matching.

\begin{figure}[h]
  \centering
  \includegraphics[width=0.45\textwidth]{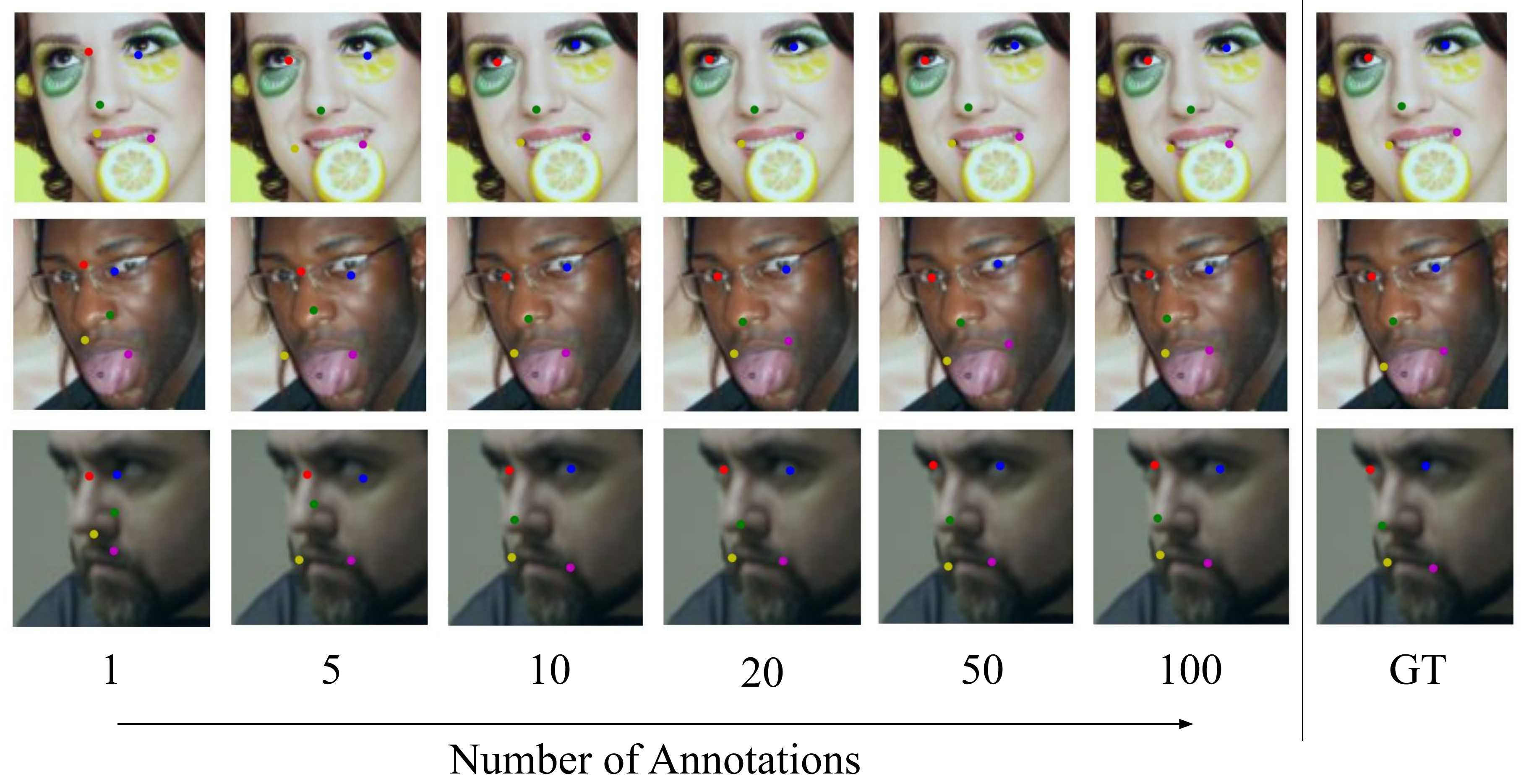}
  \caption{\textbf{Number of annotations.} Landmark prediction under different number of annotated images used for supervised training (mentioned below every column) on AFLW$_{M}$. }
  \label{fig:fewshot}
\end{figure}

\begin{figure}[h]
  \centering
  \includegraphics[width=0.48\textwidth]{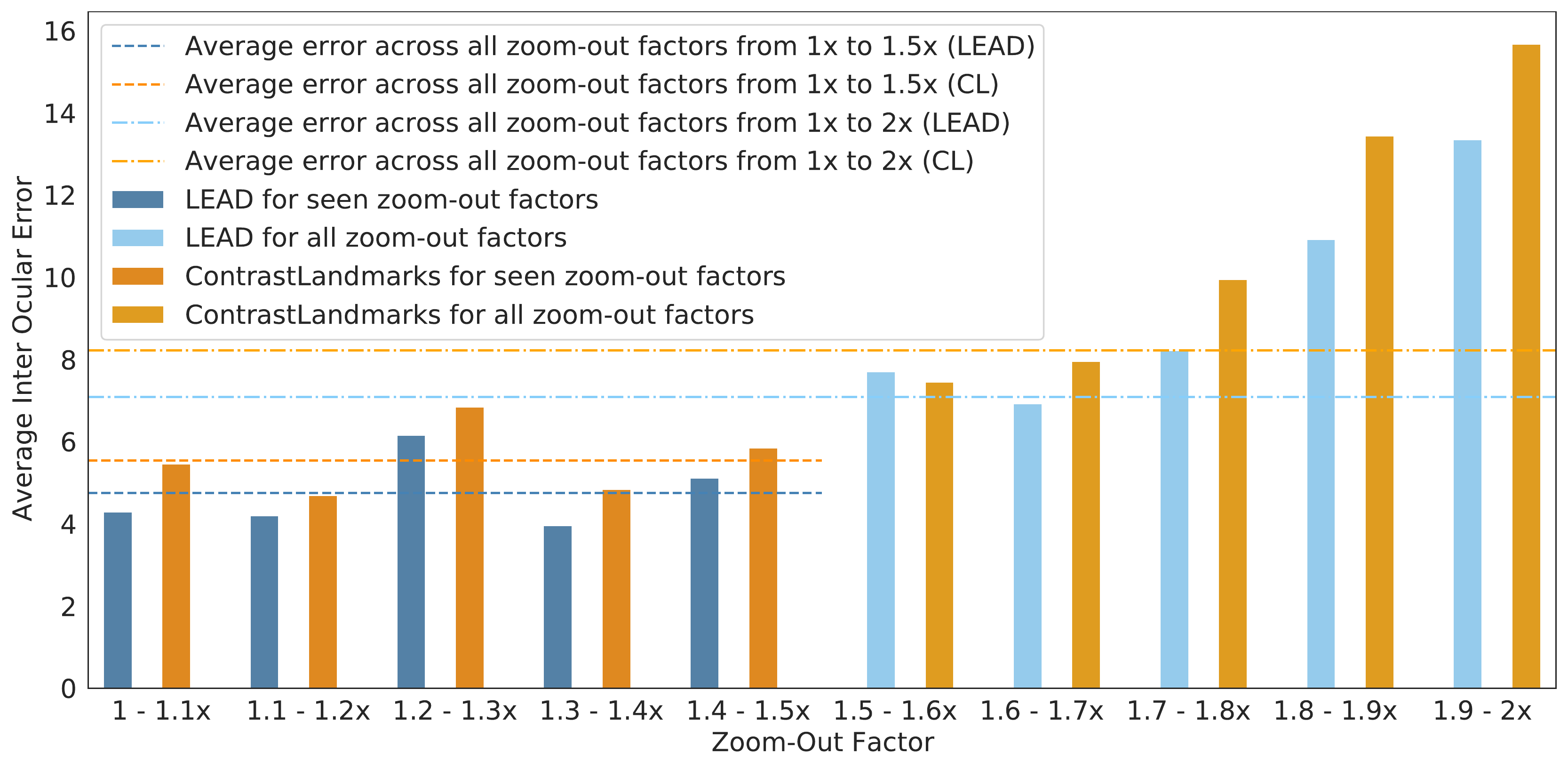}
  \caption{\textbf{Sensitivity to scale variations:} Sensitivity to seen scales (Zoom-out factor $\in$ 1-1.5x) vs unseen scales (Zoom-out factor $\in$ 1.5-2x) on unaligned-MAFL. LEAD performs better across scale changes, and is also less sensitive to \textit{unseen} scales of face.}
  \label{fig:scale_comp}
  \vspace{-4mm}
\end{figure}

\begin{figure}[h]
  \centering
  \includegraphics[width=0.45\textwidth]{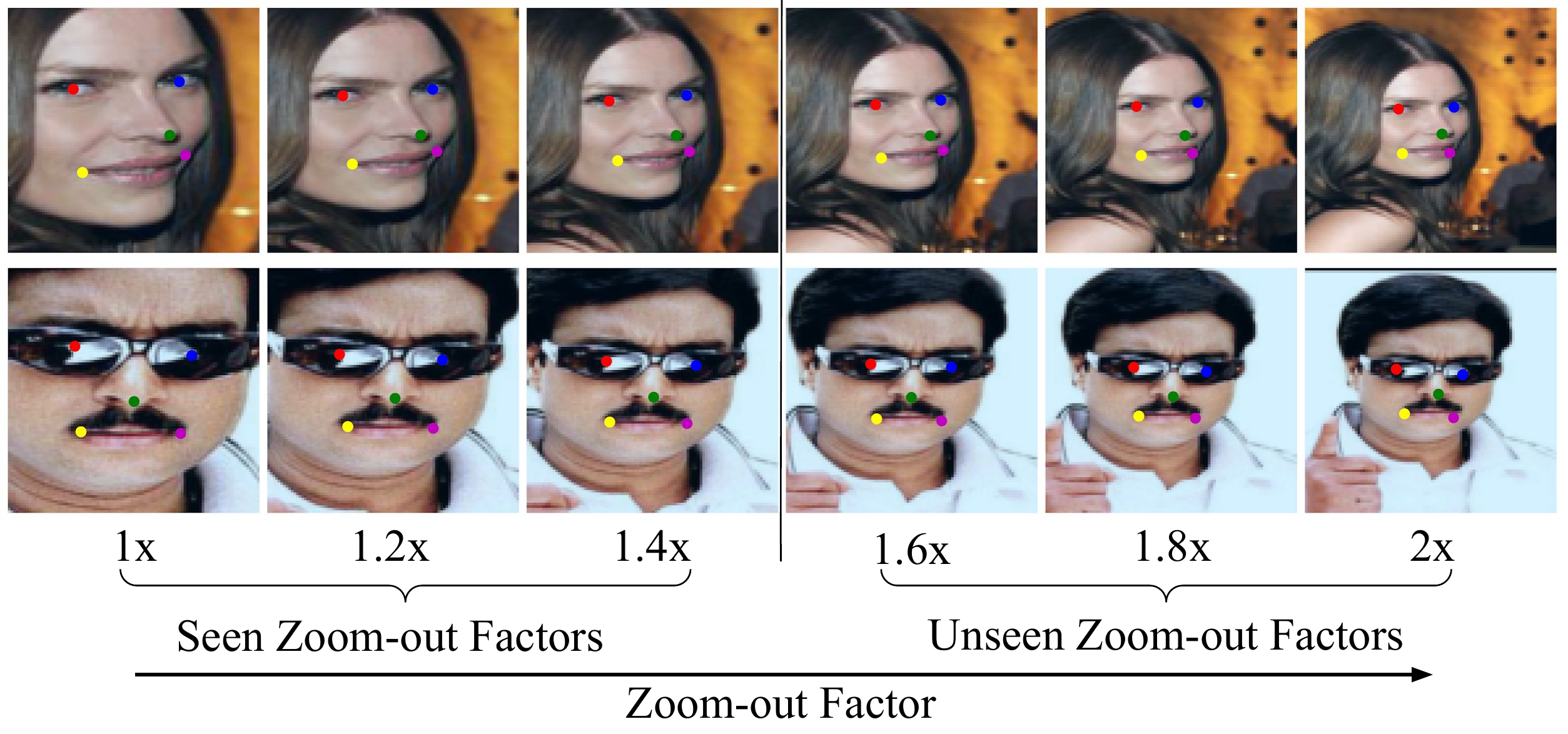}
  \caption{\textbf{Scale variation.} LEAD Landmark regression visualization across differently scaled (seen and unseen) images of unaligned-MAFL.}
  \label{fig:scale_viz}
\end{figure}

\noindent\textbf{How sensitive is it to the alignment and scale variations?} At inference stage, the landmark regressor can encounter images which may have different alignments or scales when compared to the data it was trained on. To check the sensitivity of LEAD to these changes we use features from CelebA trained LEAD to train a landmark regressor on an unaligned-MAFL dataset. We create this dataset by taking images from MAFL subset of CelebA-in-the-wild~\cite{liu2015faceattributes} dataset cropped by the bounding box annotations. Furthermore, before taking a crop, we also randomly scale up the side length of the bounding box a factor uniformly randomly sampled between 1-1.5$\times$. This results in zooming out of the image (ref. Fig.~\ref{fig:scale_viz}). We refer to this factor as ``Zoom-out factor" We evaluate the regressor on the test split which is created by scaling up the side length of the bounding box by a zoom-out factor of 1-2$\times$ before cropping. We use 64d feature for this experiment. In Fig.~\ref{fig:scale_comp}, we observe that across the range of evaluated scales, LEAD outperforms CL~\cite{cheng2020unsupervised}\footnote{Same training and evaluation protocol was followed for both.}. The gap between the two methods widens for larger zoom-out factors, which are unseen during training. We visualize the landmark regression against scale changes in Fig.~\ref{fig:scale_viz}.

\noindent\textbf{How many annotated images are required for supervised training during evaluation?}
\label{sec:anno}
 Since the evaluation of our method depends on the annotated samples, we run an ablation on the number of annotations required. We report the quantitative results in the Table~\ref{tab:fewshot}, along with qualitative annotation-wise comparisons in Fig.~\ref{fig:fewshot}. We test by varying the number of annotations to 1, 5, 10, 20, 50, and 100. We observe a consistent and significant gain in the performance with increasing number of annotations over the competent methods, a trend which even continues at different dimensions of features.
 

\section{Conclusions}
In this work, we demonstrate the superiority of the LEAD framework to learn representation at instance level from a category specific dataset. We further utilize this prior to train a dense and compact representation of the image, guided by the correspondence matching property of the learnt representation. Our experiments demonstrate the superiority of the BYOL objective over contrastive tasks like MoCo on category specific data for landmark detection. Our proposed dimensionality reduction method improves the results on both feature extractors. A future research direction could be the usage of this correspondence matching property to learn a variety of dense prediction tasks. 

\noindent
\textbf{Acknowledgements}
This work was supported by MeitY (Ministry of Electronics and Information Technology) project (No. 4(16)2019-ITEA), Govt. of India.
\medskip

\appendix
\onecolumn
\input{supple.tex}

{\small
\bibliographystyle{ieee_fullname}
\bibliography{egbib}
}

\end{document}

%% file: supple.tex







\doparttoc 
\faketableofcontents 

\part{Supplementary Materials} 


%




In this work, we present landmark prediction using our new method LEAD on a variety of face datasets. Two main assumptions in the problem setting that we presented were: 1) Availability of large-scale unannotated facial dataset for self-supervised pretraining, and 2) Availability of annotations for small scale supervised training. The reasoning behind such a setting is abundance of unannotated images available on the internet which can be leveraged to learn better features, while annotated images are hard to procure for training. Out of these two assumptions, we addressed the second one in the paper. In this work, we address the first assumption. Additionally, we examine the features from the lens of interpretability and present results on the ``bird" category.

The rest of supplementary work is organized as following: we first present the results of using in-the-wild face images for self-supervised pretraining (in Sec.~\ref{sec:itw}). Additionally, we show LEAD's efficacy on a more challenging bird landmark prediction (in Sec.~\ref{sec:cub}). We then present the interpretability of features learnt by our model by  part-discovery (in Sec.~\ref{sec:nmf}), followed by the implementation details in Sec.~\ref{sec:impl}. Lastly we discuss interpetability of denser intermediate outputs of our model by clustering (in Sec.~\ref{sec:intr}), followed by some additional visuals for the scale ablation as performed in the main paper (Sec.~\ref{sec:add_res}).

\section{In-the-wild pretraining}
\label{sec:itw}

\begin{figure}[h]
  \centering
  \includegraphics[width=\textwidth]{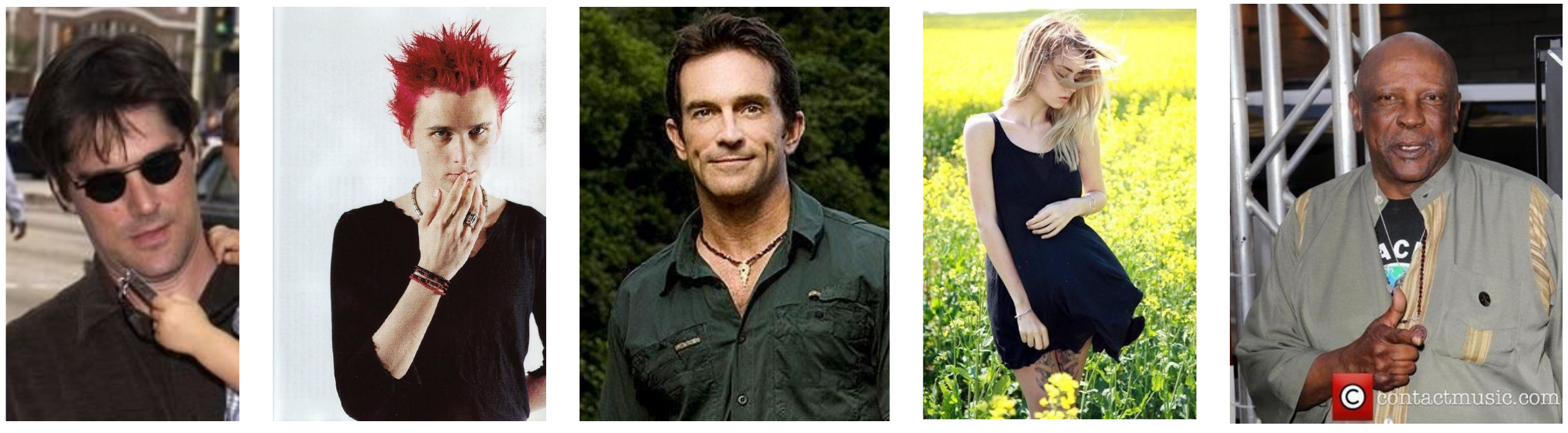}
  \caption{Samples from CelebA In-the-wild}
  \label{fig:celebAitw}
\end{figure}

While we use large-scale unannotated dataset for pretraining, it is still a well-cropped data where the training images were completely occupied by the face. In this experiment, we use uncropped in-the-wild images from the CelebA~\cite{liu2015faceattributes} dataset. Samples from the in-the-wild CelebA dataset are shown in fig.~\ref{fig:celebAitw}.

We refer the reader to table~\ref{tab:itw} for the quantitative results in this setting. The pretraining on this dataset proves to be beneficial in some evaluation datasets, while still giving improvements over others. Pretraining on CelebA In-the-wild shows impressive results on AFLW$_R$ and 300W, which are better than those obtained by CelebA pretraining. For the remaining 2 datasets, the inter-ocular distance slightly increases compared to that of celebA pretraining, but is still better than the prior arts. This performance can be reasoned as the in-the-wild images have large variations between different regions in the image. As every time when the crops of image are taken as augmentation, each crop provides the network with a variety of image distributions and statistics to learn from. While in case of `cropped celebA' pretraining, a large area of the image is occupied face, hence there is less variety in the kinds of crops it results in. 

\begin{figure}[h]
  \centering
  \includegraphics[width=\textwidth]{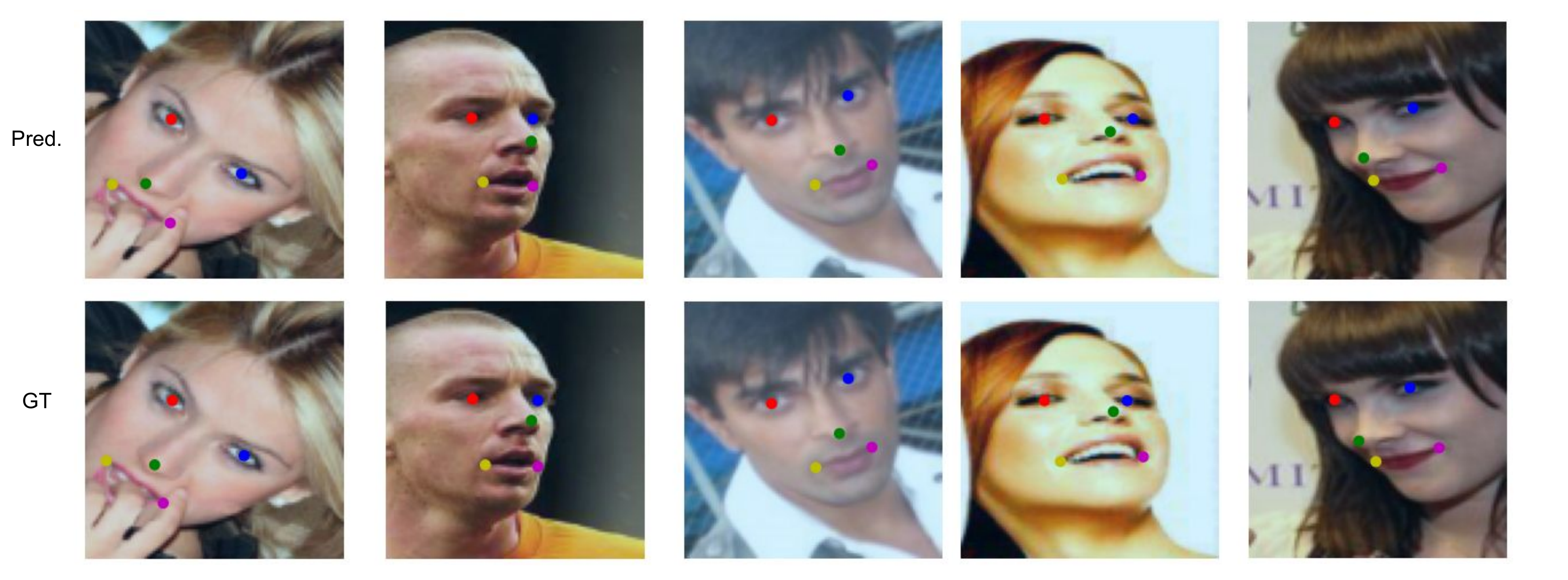}
  \caption{Landmark prediction on unaligned MAFL (in-the-wild)}
  \label{fig:maflitw}
\end{figure}

We also evaluate this model on unaligned face images. These images use the same split as MAFL, but from CelebA in-the-wild. We call is MAFL in-the-wild. For evaluation in this setting we randomly sample a percentage between $10$ to $20$ to increase the height and width of the original face bounding box for each image, this bigger bounding box is then used to crop the respective images. 
This helps in breaking the alignment of faces, that is present in the cropped MAFL dataset. We report the results on this evaluation in Table~\ref{tab:maflitw} and Fig.~\ref{fig:maflitw}. The inter-ocular distance on this dataset is compared with both DVE and ContrastLandmarks. Our method gives \textit{9.75\%} and \textit{2.88\%} better relative IOD then DVE and ContrastLandmarks respectively.
\begin{table}[h]
\parbox{.55\textwidth}{
\centering
\caption{Effect of pretraining on in-the-wild face dataset. }
\label{tab:itw}
\begin{tabular}{llllll}
\hline
\multicolumn{1}{c}{\textbf{Method}} & \textbf{Feat. dim} & \multicolumn{1}{c}{\textbf{MAFL}} & \multicolumn{1}{c}{\textbf{AFLW}$_M$} & \textbf{AFLW}$_R$ & \textbf{300W} \\ \hline
DVE~\cite{Thewlis19a}                                 & 64                 & 3.23                              & 8.52                              & 7.38          & \textbf{5.05} \\
CL~\cite{cheng2020unsupervised}                                  & 64                 & \textbf{3.00}                     & 7.87                              & 6.92          & 5.59          \\
LEAD (ours)                                 & 64                 & 3.01                              & \textbf{6.81}                     & \textbf{6.42} & 5.34          \\ \hline
CL~\cite{cheng2020unsupervised}                                    & 128                & \textbf{2.88}                     & 7.81                              & 6.79          & \textbf{5.37} \\
LEAD (ours)                                 & 128                & 3.07                              & \textbf{6.79}                     & \textbf{6.38} & 5.56          \\ \hline
CL~\cite{cheng2020unsupervised}                                  & 256                & \textbf{2.82}                     & 7.69                              & 6.67          & \textbf{5.27} \\
LEAD (ours)                                 & 256                & 3.09                              & \textbf{6.85}                     & \textbf{6.22} & 5.53          \\ \hline
CL~\cite{cheng2020unsupervised}                                  & 3840               & \textbf{2.46}                              & 7.57                              & 6.29          & 5.04          \\
LEAD (ours)                                 & 3840               & \textbf{2.46}                     & \textbf{6.48}                    & \textbf{5.64} & \textbf{4.47}
\\ \hline
\end{tabular}
}
\hfill 
\parbox{.35\textwidth}{
\centering
\caption{Evaluation on MAFL in-the-wild. }
\label{tab:maflitw}
\begin{tabular}{lll}
\hline
\textbf{Method} & \textbf{Feat. Dim.} & \textbf{IOD}  \\ \hline
DVE~\cite{Thewlis19a}              & 64                & 4.51          \\
LEAD (Ours)     & 64                & \textbf{4.07} \\ \hline
CL~\cite{cheng2020unsupervised}              & 3840                & 3.12          \\
LEAD (Ours)     & 3840                & \textbf{3.03} \\ \hline
\end{tabular}
}
\end{table}

\vspace{6mm}

\section{Bird Landmark Prediction}
\label{sec:cub}
\begin{wraptable}{r}{4.75cm}
\centering
\caption{PCK (Percentage of correct keypoints) on CUB dataset}
\label{tab:cub}
\begin{tabular}{lll}
\hline
\textbf{Method} & \textbf{Feat. Dim.} & \textbf{PCK}  \\ \hline
CL\cite{cheng2020unsupervised}              & 3840                & 68.63         \\
LEAD (Ours)     & 3840                & 67.31 \\ \hline
\end{tabular}
\end{wraptable}

\begin{figure}[h]
  \centering
  \includegraphics[width=0.8\textwidth]{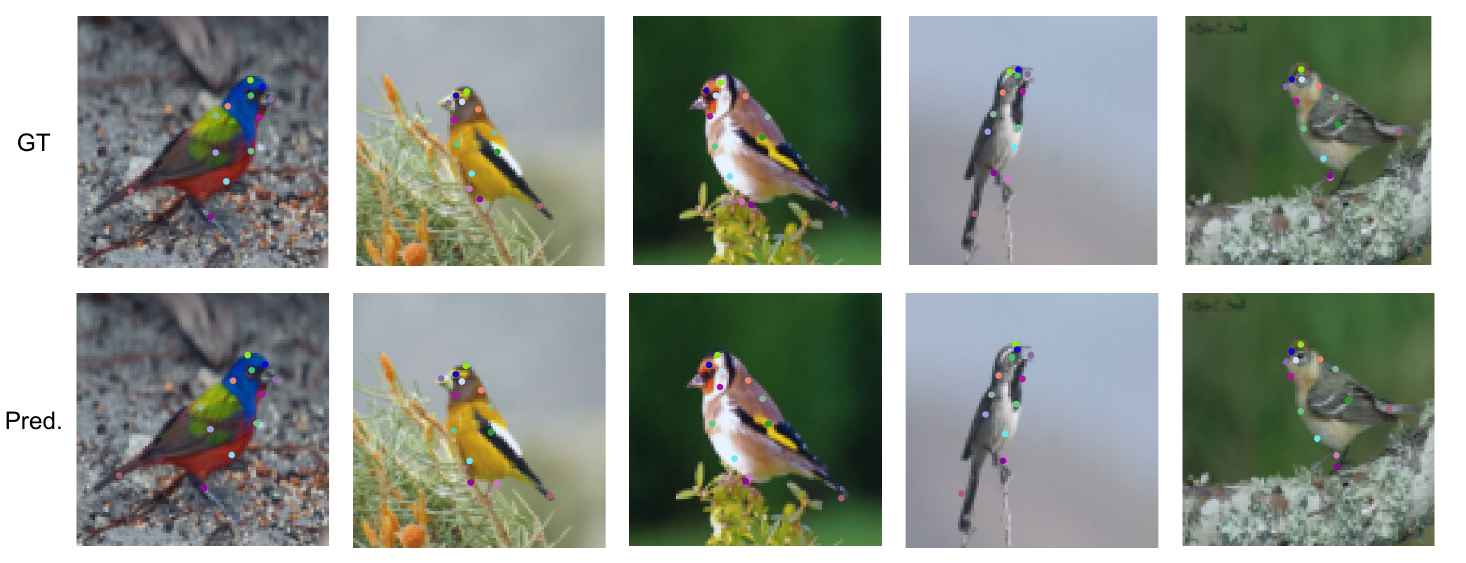}
  \caption{Landmark prediction on birds from CUB dataset}
  \label{fig:cub}
\end{figure}

We demonstrate the efficacy of our method on challenging bird landmark prediction task. We show the results in Fig.~\ref{fig:cub}. In this task, we train the instance-level model on iNat17-Aves dataset, which contains in-the-wild images of ``Aves" class from iNaturalist 2017 dataset ~\cite{inat}. This is followed by supervised training on a subset of CUB~\cite{WahCUB_200_2011} dataset, containing 35 species from Passeroidea super-family. CUB images have 15 annotated landmarks. For both the datasets, we use the same split as ~\cite{cheng2020unsupervised}. We compare the percentage of correct keypoints. For PCK computation, a prediction is considered to be correct if its distance from the ground-truth keypoint is within 5\% of the longer side of the image. Occluded keypoints are ignored during the evaluation. We report a competitive PCK of 67.3\%, which we compare against ContrastLandmarks~\cite{cheng2020unsupervised} in Table~\ref{tab:cub}.

\section{Part Discovery}
\label{sec:nmf}
\begin{figure}[h]
    \centering
    \includegraphics[width=0.8\textwidth]{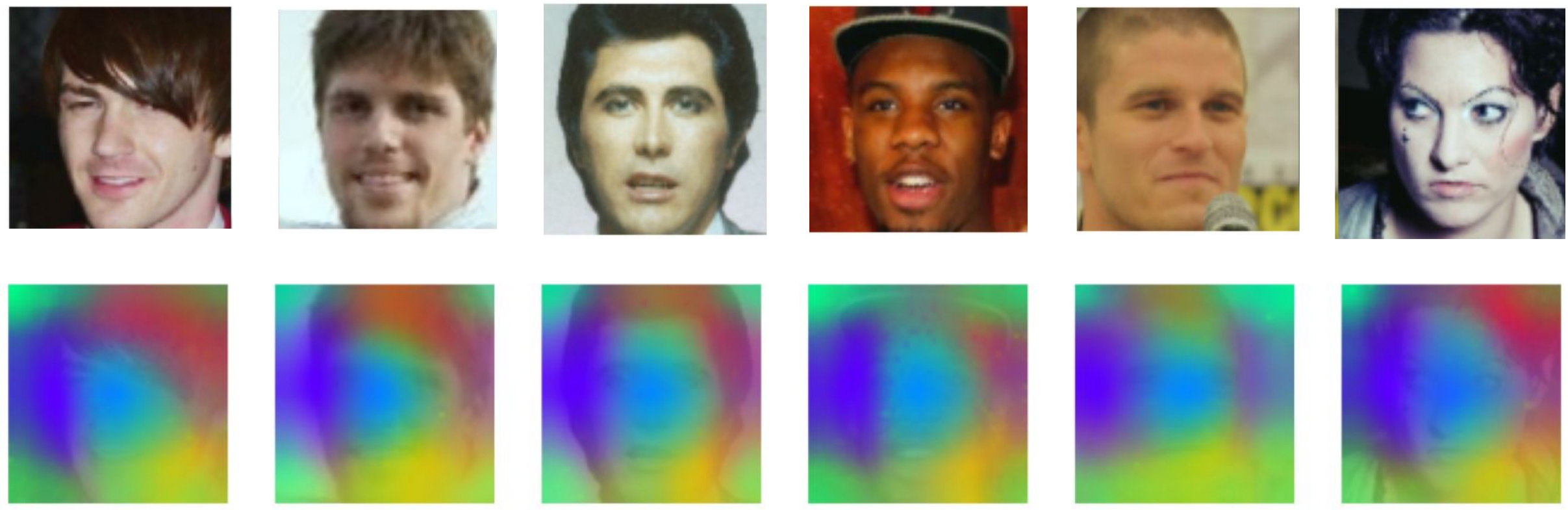}
    \caption{NMF part clustering~\cite{collins2018} of learnt embeddings. Each color represents a discovered part.}
    \label{fig:nmf}
\end{figure}

We perform deep feature factorization~\cite{collins2018} on the representation obtained from stage 1 of training for part discovery. It can be noted in Fig.~\ref{fig:nmf} that parts discovered by this method are consistent across instances. Intensity of the color denotes the presence of the part. It is interesting to observe the intensity of the color corresponding to the hair is very less in case of the fourth and fifth instance in Fig.~\ref{fig:nmf}, where the person is wearing a cap and has less hair respectively.

\section{Implementation details}
\label{sec:impl}

\textbf{Stage 1: Instance-level training}. We use BYOL~\cite{grill2020bootstrap} to train the unsupervised representations. We train BYOL for 200 epochs with a batch size of 256 and use a cosine learning rate scheduler \cite{Loshchilov2017SGDRSG}, with a warm-up period of 2 epochs. We follow the the augmentation pipeline as proposed in BYOL, wherein we use solarization as an augmentation only for the target encoder. We use the publicly available BYOL implementation from OpenSelfSup\footnote{\href{https://github.com/open-mmlab/OpenSelfSup}{https://github.com/open-mmlab/OpenSelfSup}}. For comparison with DVE \cite{Thewlis19a} and ContrastLandmarks \cite{cheng2020unsupervised}, we use the released pretrained models from the respective official code repositories.

\textbf{Stage 2: Dense training}. Here we train an FPN decoder \cite{fpn} while keeping the learned backbone encoder to be frozen. We train the decoder for 10 epochs with a batch size of 256, using a cosine learning rate scheduler with a warm-up of 2 epochs, similar to stage 1. We again follow the BYOL augmentation pipeline for training. We set the temperature $\tau$ to be 0.05 (Refer to Eq. 2 in the main paper).

\textbf{Training of Supervised Landmark Regressors.} For a fairer comparison we train the supervised regressors with frozen feature extractor exactly as proposed in \cite{cheng2020unsupervised}. We also show a comparison of supervised training speeds against different feature dimensions in Table~\ref{tab:speed}.


\begin{wraptable}{r}{5.5cm}
\centering
\label{tab:speed}
\caption{Comparison of supervised training speeds at differ feature dimensions. Note that hypercolumn features (3840 feat. dim.) are 55$\times$ slower.}
\begin{tabular}{ll}
\hline
Feat. Dim. & FLOPS \\ \hline
3840       & 2.21  \\
256        & 0.16  \\
128        & 0.08  \\
64         & 0.04  \\ \hline
\end{tabular}
\end{wraptable}

\section{Interpretability Analysis}
\label{sec:intr}
\begin{figure}
    \centering
    \includegraphics[width=0.85\textwidth]{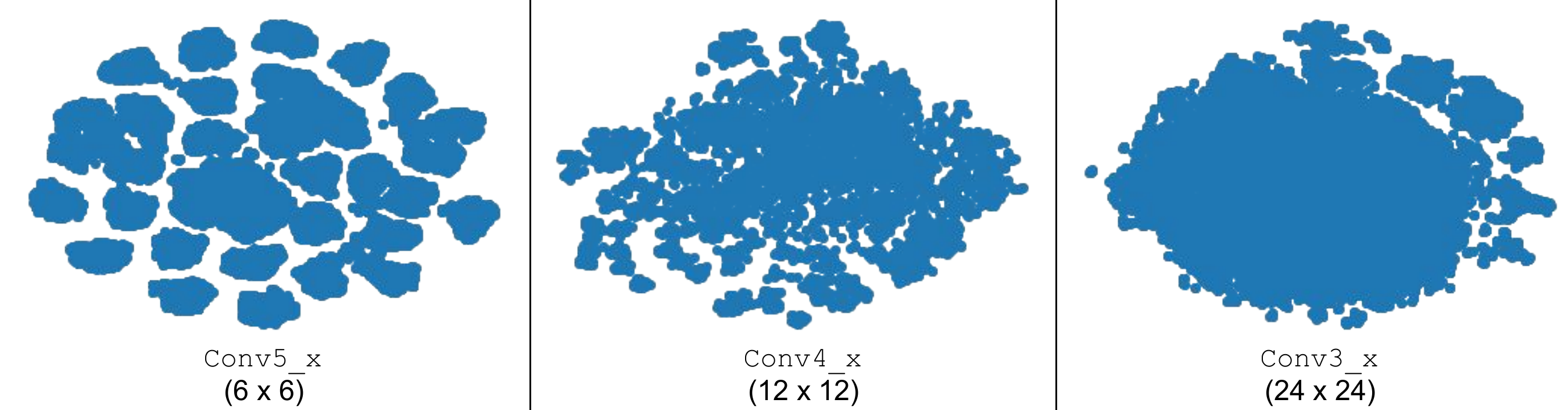}
    \caption{t-SNE plots of the intermediate layers' feature maps obtained after training (stage 1 of our model). Spatial dimension of the feature map indicated in brackets.}
    \label{fig:byol_clusters_6x6}
\end{figure}

\begin{figure}
    \centering
    \includegraphics[width=\textwidth]{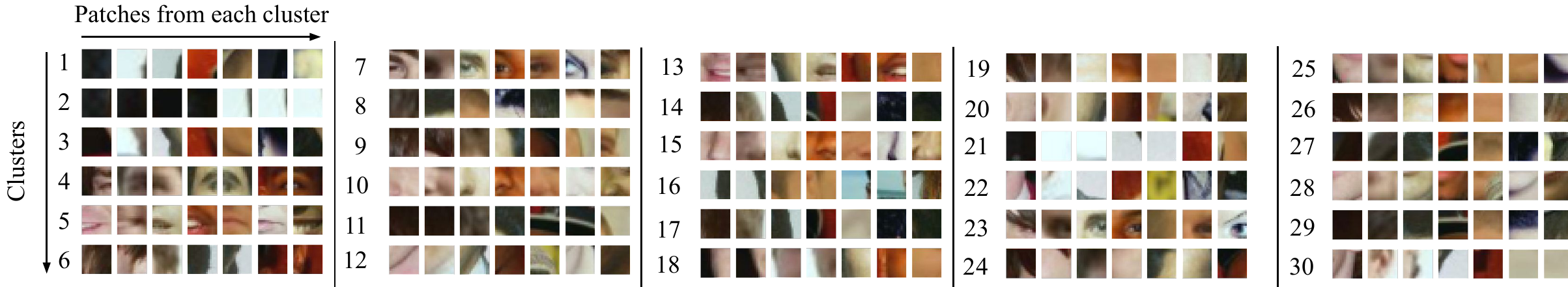}
    \caption{t-SNE embeddings tend to cluster part-wise. Each row shows patches from a cluster obtained from higher spatial resolution features (6$\times$6 in this case). The 30 clusters shown corresponds to 30 clusters in Fig.~\ref{fig:byol_clusters_6x6}. Each row denotes a cluster, which contains patch of semantically meaningful part of face. (Mouth: clusters 5, 13; Left and right eye: cluster 7 and 23, Nose: clusters 10, 15; Left and right jaw: clusters 18 and 12. Other clusters contain less discriminative parts like cheeks and forehead.)}
    \label{fig:tsne_6x6}
\end{figure}

We further examine the interpretabilty of the t-SNE embeddings obtained by clustering intermediate representations of the resnet model thereby capturing denser grids ($6 \times 6, 12 \times 12 $ and $24 \times 24$).  Fig.~\ref{fig:byol_clusters_6x6} shows emergence of semantically meaningful clusters in the $6 \times 6$ grid, Fig.~\ref{fig:tsne_6x6} shows the corresponding parts captured by these clusters. It can also be observed that for further denser grids ($12 \times 12 $ and $24 \times 24$) we see the emergence of one big cluster which cannot be easily split into semantically meaningful regions. We further observed emergence of a single big cluster even after stage-2 training of our method, the possible cause of which may be the reduced dimensionality  after stage 2 hurting the expressiveness as t-SNE embedding compared to the original hypercolumn ($3840D$).


\section{Additional Visuals for Scale Ablations}
\label{sec:add_res}
\begin{figure}
    \centering
    \includegraphics[width=0.925\textwidth]{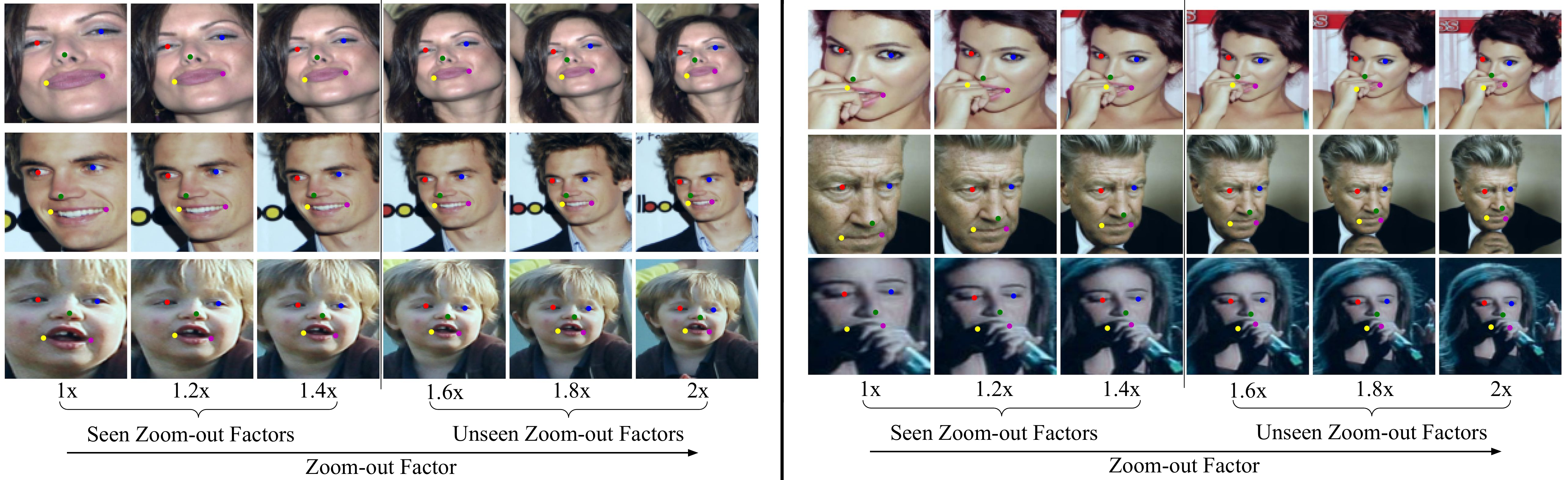}
    \caption{Additional Results on Scale ablation on seen scale (zoom-out factor $\in$ 1-1.5x) and unseen scale (zoom-out factor $\in$ 1.5-2x) variations of the faces from unaligned-MAFL dataset}
    \label{fig:scale_viz_1}
\end{figure}

We present additional examples of generalization of LEAD to seen and unseen scales of input face. These are shown in Fig.~\ref{fig:scale_viz_1}



\clearpage